\def\year{2021}\relax
\documentclass[letterpaper]{article} 
\usepackage{aaai21}  
\usepackage{times}  
\usepackage{helvet} 
\usepackage{courier}  
\usepackage[hyphens]{url}  
\usepackage{graphicx} 
\usepackage{natbib}  
\usepackage{caption} 

\usepackage{array,multirow}
\usepackage{enumitem}
\usepackage{graphicx}
\usepackage{booktabs}
\usepackage{amsmath}
\usepackage{amsfonts}
\usepackage{xcolor}
\usepackage{verbatim}
\usepackage{subcaption} 
\graphicspath{ {figs/} }

\newcommand{\newcite}[1]{\citeauthor{#1} \shortcite{#1}}

\newcommand{\newdatasetname}[1]{\textsc{QmdsCnn}}
\newcommand{\irdatasetname}[1]{\textsc{QmdsIr}}
\newcommand{\jointdatasetname}[1]{\textsc{QmdsCnnIr}}
\newcommand{\ourmodel}[1]{HEROSumm}

\title{Data Augmentation for Abstractive Query-Focused \\ Multi-Document Summarization}
\author{
Ramakanth Pasunuru,\textsuperscript{\rm 1}
Asli Celikyilmaz,\textsuperscript{\rm 2}
Michel Galley,\textsuperscript{\rm 2}
Chenyan Xiong,\textsuperscript{\rm 2}\\
Yizhe Zhang,\textsuperscript{\rm 2}
Mohit Bansal,\textsuperscript{\rm 1}
Jianfeng Gao\textsuperscript{\rm 2} \\
}
\affiliations{
    \textsuperscript{\rm 1}UNC Chapel Hill,
    \textsuperscript{\rm 2}Microsoft Research, Redmond \\
    \{ram, mbansal\}@cs.unc.edu,
    \{aslicel, mgalley, Chenyan.Xiong, yizhe.zhang, jfgao\}@microsoft.com

}

\begin{document}

\maketitle

\begin{abstract}
The progress in Query-focused Multi-Document Summarization (QMDS) has been limited by the lack of sufficient large-scale high-quality training datasets. 
We present two QMDS training datasets, which we construct using two data augmentation methods: (1)~transferring the commonly used single-document CNN/Daily Mail summarization dataset to create the \newdatasetname{} dataset, and (2) mining search-query logs to create the \irdatasetname{} dataset. These two datasets have complementary properties, i.e., \newdatasetname{} has real summaries but queries are simulated, while \irdatasetname{} has real queries but simulated summaries.
To cover both these real summary and query aspects, we build abstractive end-to-end neural network models on the combined datasets that yield new state-of-the-art transfer results on DUC datasets.
We also introduce new hierarchical encoders that enable a more efficient encoding of the query together with multiple documents. 
Empirical results demonstrate that
our data augmentation and encoding methods outperform baseline models on automatic metrics, as well as on human evaluations along multiple attributes.\footnote{Code: \url{https://github.com/ramakanth-pasunuru/QmdsCnnIr}}

\end{abstract}

\begin{figure}[t]
\centering
\includegraphics[width=0.98\linewidth]{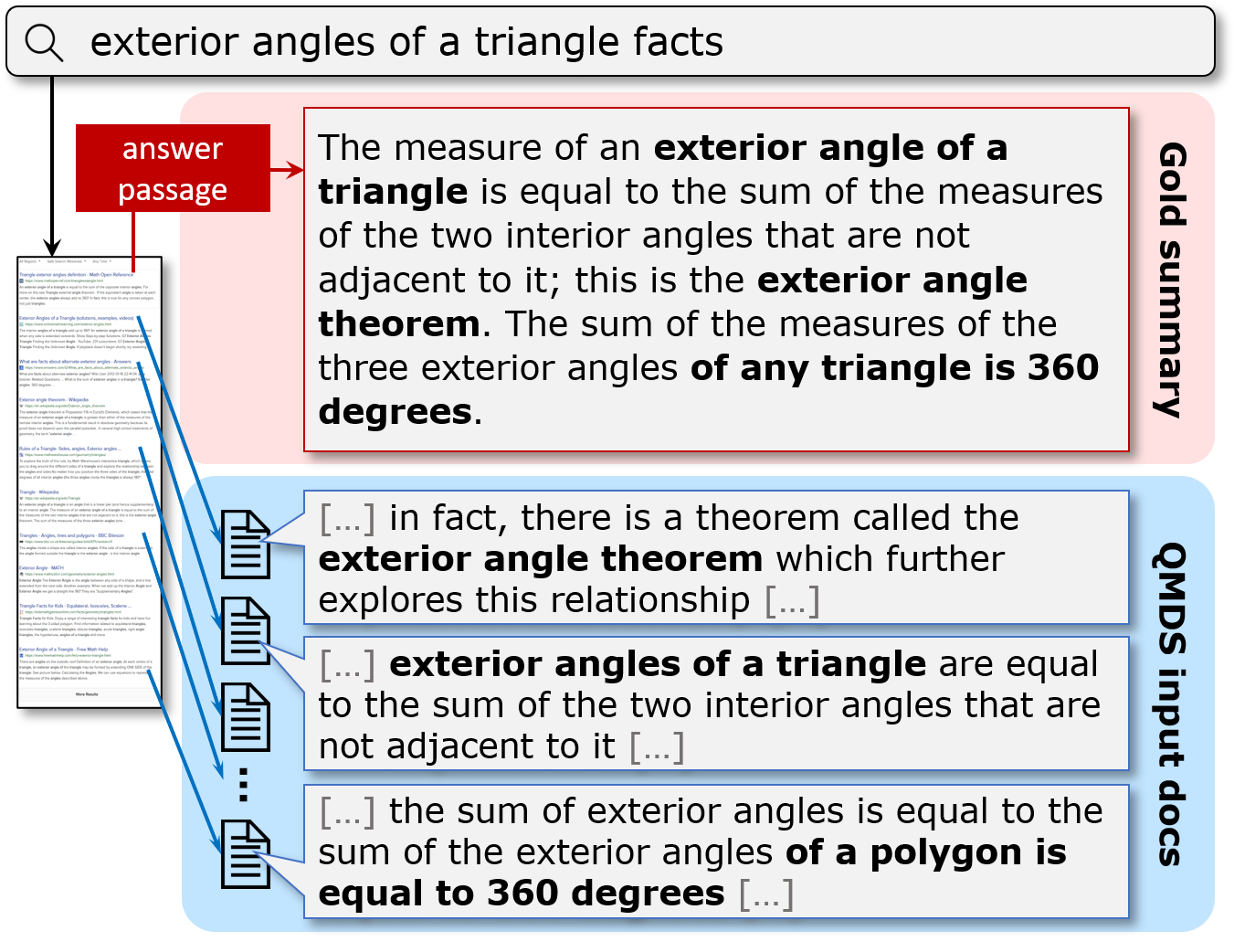}
\caption{Sample from our \irdatasetname{} dataset, which illustrates how a set of retrieved documents based on a query can provide complementary information that can be used to reconstruct the information in the answer passage (used as gold summary).
}
\label{fig:irdata}
\end{figure}

\section{Introduction}
Query-focused multi-document summarization (QMDS) aims at generating a short summary from a set of documents that answers a query. Compared to the popular single document summarization (SDS) task~\cite{rush2015neural,chopra2016abstractive,nallapati2016abstractive,celikyilmaz2018deep,chen2018fast,gehrmann2018bottom}, research in QMDS has received less attention. This is partially due to the scarcity of large-scale high-quality QMDS datasets. The SDS has variety of high-quality datasets~\cite{hermann2015teaching,grusky2018newsroom} on different domains such as news~\citep{giga,nallapati2016abstractive}, scientific articles \citep{qazvinian-radev-2008-scientific}, etc., however, not many high-quality datasets exist for QMDS training and evaluation. 

Another overlooked feature of QMDS is that it tries to solve more realistic query-based scenarios than the SDS or multi-document summarization (MDS) tasks. Different from these tasks, QMDS task considers summarizing only salient information that best answers the query in a logical order. In fact, QMDS is more realistic for various applications such as personalized information retrieval (IR), conversational IR, and recommendation engines, in which search results can be tailored to an information need. To support research on challenging QMDS task, we introduce two data augmentation methods and new neural models.  

\noindent\textbf{Two new data augmentation methods.}
Recently, multiple new MDS datasets have been introduced: A large-scale dataset by ~\newcite{liu2018generating} named WikiSum, and a smaller MDS by ~\newcite{fabbri2019multi}. Even though WikiSum also includes the topic of the article as query, it is mostly used to train MDS models~\cite{liu2018generating,liu2019hierarchical}, since a topic is more generic to be used as an information seeking query.
To this end, we introduce two new data augmentation methods to enable large-scale training of QMDS models.
In the first method, we restructure the single-document CNN/Daily Mail (CNN/DM) dataset~\cite{hermann2015teaching} to create a new QMDS dataset by chunking the documents into small documents of paragraphs and use the title of the article as a query. We refer to this dataset as \newdatasetname{}, which has $\sim$300K samples.
For the second method, we mine real-user web queries, top ranked web documents, and answer passages from the search log of Bing (see Fig.~\ref{fig:irdata}). We consider the answer passage returned by Bing, which is extracted from one of the top ranked documents as the summary and the rest of the documents as input documents forming our second QMDS dataset. We call this dataset as \irdatasetname{}, which has $\sim$100K samples. These two new datasets have complementary properties: \newdatasetname{} has manually written summaries and noisy queries, while the \irdatasetname{} has real queries but automatically generated summaries. Thus, we combine these two datasets to obtain a balanced set of high-quality augmented data, which we used to train our novel, large-scale QMDS models.

\noindent\textbf{Novel models for query-focused MDS task.} ~\newcite{liu2019hierarchical} presented a hierarchical encoder-decoder transformer MDS model with attention layers and incorporated the query by simply concatenating to the top-ranked document. 
Focusing on building a better abstractive end-to-end neural network-based QMDS model, we introduce \textbf{\ourmodel{}}:
\textbf{H}i\textbf{E}rarchical Que\textbf{R}y focused \textbf{O}rder-aware multi-document \textbf{Summ}arization model, extending the model in~\newcite{liu2019hierarchical} with three novel components:
(a) \textbf{\emph{Hierarchical Encoding}}: unlike previous work, which uses a single global representation of the multi-document encoder during the decoding of the summary, we use both the local and global representations from the encoder during decoding; (b) \textbf{\emph{Ordering Component}}: 
The QMDS model of~\cite{liu2019hierarchical} receives the rank order of documents as input from an external module. If the order information is incorrect, it can adversely affect the QMDS model's performance.
Hence, to eliminate this cascading error effect, we introduce a new document ordering module that learns the ordering pattern while training the QMDS model parameters end-to-end. (c) \textbf{\emph{Query Component}}: 
Unlike previous work, which prepends the query to top document during encoding, we enrich our QMDS model with an additional transformer component that encodes the query. The decoder then attends the local and/or global layers of the multiple document encoders which are conditioned on the query output encoding.

Our quantitative evaluations show that the \ourmodel{} model, which includes new QMDS focused components, can generate more accurate summaries than the baseline. 
We also demonstrate that neural models trained on the \newdatasetname{} and \irdatasetname{} datasets constructed with our data augmentation methods show promising attributes of transferability compared to the models trained on the WikiSum dataset, when tested on real QMDS datasets with summaries written by humans (DUC 2006 and 2007).
We further validate the superiority of our data augmentation methods via human evaluation studies along multiple attributes.


\section{Related Work}
\paragraph{Earlier MDS Research.}
Earlier extractive MDS work have used various approaches including maximum marginal relevance (MMR) to reduce redundancy~\cite{carbonell1998use}, clustering based on topic detection~\cite{radev2004centroid}, graph-based~\cite{erkan2004lexrank} or hierarchical LDA-style models~\cite{haghighi2009exploring}, and variants of query-focused summarization~\cite{dang2005overview}, that orient the summary around a given query~\cite{daume2006bayesian,zhao2009using}. Earlier abstractive MDS focused on template- and planner-based~\cite{mckeown1995generating,radev1998generating,barzilay1999information} and graph-based methods~\cite{ganesan2010opinosis}.

\noindent\textbf{Recent MDS Research.}
Recent neural SDS models have shown significant improvements on both extractive~\cite{nallapati2016classify,cheng2016neural,narayan2018ranking} and abstractive~\cite{rush2015neural,chopra2016abstractive,nallapati2016abstractive,celikyilmaz2018deep,chen2018fast,gehrmann2018bottom} setups. However, MDS models with neural networks are limited by the unavailability of large-scale MDS datasets.~\newcite{zhang2018adapting} adapts a state-of-the-art SDS model for MDS.~\citet{feigenblat2017unsupervised} introduces a extractive-based QMDS model using the Cross-Entropy method.~\newcite{baumel2018query} introduces query relevance to adapt SDS model to QMDS.~\newcite{lebanoff2018adapting} exploits the MMR method to fuse disparate sentences from multi-document inputs.~\newcite{liu2019hierarchical} introduced a hierarchical transformer model to better encode global and local aspects in multiple documents. 
In this work, focusing on the \textit{coherency} aspect of summaries, we design components to attend the query and the local and global aspects of documents better, while tracking the \textit{ordering} information to generate more accurate and focused summaries.

\noindent\textbf{MDS Datasets.}
Recent introduction of large-scale MDS datasets, WikiSum ~\citep{liu2018generating}, Multi-News~\citep{fabbri2019multi}, Wikipedia Current Events~\citep{ghal2020largescale}, set a promising direction for developing powerful neural network models.
Focusing on \textit{query} focused summarization of more realistic scenarios, we constructed two new large-scale QMDS datasets, based on popular single document CNN/DM dataset and real search query logs.

\section{Two New QMDS Datasets}
\label{sec:new-qmds-datasets}

\subsection{\newdatasetname{} Dataset}
\label{sec:cnndm-qmds-dataset}
CNN/Daily Mail (CNN/DM) is a commonly used SDS dataset \citep{hermann2015teaching}. The documents are online news articles and the summaries are human written highlights of corresponding articles. We use the scripts provided by~\newcite{see2017get} to obtain this dataset. We present here step-by-step instructions for converting the CNN/DM SDS dataset into a QMDS dataset:

\noindent \textbf{Step-1: Generate a query per document.} We use the title of news article as the query to enable a query-focused setup. 

\noindent \textbf{Step-2: Chunk documents.}
Each news article has multiple small paragraphs (approximately 20 small paragraphs), and the sentences in the summary span across these small paragraphs. We randomly group these small paragraphs into chunks (one to four paragraphs per chunk), each chunk forming a new document. In essence, we split the original article into anywhere from one to four smaller documents, composing new \textbf{query-documents-summary triplets}. 

\noindent \textbf{Step-3: Create new documents from documents on the same topic.}
CNN/DM dataset contains several documents on similar or the same topic, mostly written by different newsgroups on the same day. 
Our goal is to collate these documents of similar topics and select chunks from them to append to our new triplets datasets as follows:\footnote{In the scenario where there are no similar topics, the retrieved documents are still useful to simulate the use case of QMDS for presenting the search results, where the returned document set contains both relevant and irrelevant documents.}
We take the entire CNN/DM dataset and index all the chunks with BM25~\cite{robertson1994some}.\footnote{BM25 is a ranking function used by search engines to estimate the relevance of documents to a given search query.} For each newly constructed query-documents-summary triplet, we take the title of the summary as query, send to the BM25 search engine, which returns chunks from the entire dataset related to the title (as query). 
We take the top four chunks and append to the original query-documents-summary triplet as new documents.
We provide details on the data curation pipeline with example triplets in the Appendix.

\begin{table}
\begin{center}
\small
\centering
\begin{tabular}{l|c|c|c}
\toprule
Statistics & Train & Val & Test  \\
\midrule
{\bf \newdatasetname{}}
(\# samples) & 287,113 & 13,368 & 11,490 \\
- Avg. \# documents & 6.5 & 6.5 & 6.5 \\
- Avg. Doc. length (\# tokens) & 355 & 346 & 353 \\
- Avg. Query length (\# tokens) & 13.8 & 14.5 & 14.2 \\
\midrule
{\bf \irdatasetname{}}
(\# samples) & 82,076 & 10,259 & 10,260 \\
- Avg. \# documents & 5.8 & 5.4 & 5.5 \\
- Avg. Doc. length (\# tokens) & 1,291 & 1,402 & 1,379 \\
- Avg. Query length (\# tokens) & 6.2 & 6.2 & 6.2\\ 
\bottomrule
\end{tabular}
\end{center}
\caption{\newdatasetname{} and \irdatasetname{} statistics.
}
\label{table:cnndm-qmds-stats}
\end{table}

Table~\ref{table:cnndm-qmds-stats} presents the statistics of \newdatasetname{} dataset. The average number of documents and document length are roughly same across train/val/test sets. Each triplet sample contains around 5-8 documents, from which four documents are retrieved using BM25 as described previously.

\noindent\textbf{Is \newdatasetname{} dataset suitable for QMDS?} 
Firstly, an accurate abstractive summary should be entailed by the input document and contain only the salient information \citep{guo-etal-2018-soft}. 
Specifically for the QMDS task, the query should also be entailed by the summary. Since the documents along with their titles and summaries are all written by humans in the CNN/DM dataset, we assume that the summaries should reflect the title, as well as each summary should be entailed by its corresponding document. We extend the document list of a given query-documents-summary triplet with additional chunks as new relevant documents. Since these relevant documents are retrieved based on the relatedness to the query (title of the summary), they extend the entailment chain such that the abstractive summary of a triplet is also entailed by the corresponding augmented documents.

Secondly, a good summary should contain sentences that span across multiple documents. We measure this by taking a summary and corresponding documents from a query-documents-summary triplet (excluding the retrieved documents), and align each sentence in the summary to one of the documents. 
We found that there are many triplets whose summary spans multiple documents, thus, enabling multi-document properties.
Statistics and additional analysis are in the Appendix.

\subsection{\irdatasetname{} Dataset}
\label{subsec:ms-dataset-description}
The \irdatasetname{} contains queries that are issued by actual search engine users. This is more realistic than using the titles of articles as queries as in the WikiSum dataset. We follow these steps to construct the \irdatasetname{} dataset:

\noindent \textbf{Step-1: Sample search logs.}  We randomly sample English queries from Bing search logs in the United States, during the first six months of 2019. Only queries that have natural language answers returned and the answer passages that received positive user feedback are kept.

\noindent \textbf{Step-2: Capture summary text and documents.} For each posed-query, we collect the top 10 ranked documents from Bing and the displayed answer passage. The answer passage is extracted from one of the top ranked documents by Bing's production QA system, which is a constantly updated state-of-the-art neural-based single-document extractive summarization model.
We use this answer passage as the target summary.
We identify the document from which the answer passage is extracted, and omit that document from the candidate documents to enforce the needs of MDS.  

\noindent \textbf{Step-4: Construct dataset.} The query, the extracted answer passage as summary, and the rest of the top-ranked documents represent the triplets of our \irdatasetname{} dataset (see Table~\ref{table:cnndm-qmds-stats} for statistics).

\noindent\textbf{Is \irdatasetname{} dataset suitable for QMDS?}
Since we use real search query logs, the documents in a triplet are closely related to the query with a potential to answer the query, however, they may or may not contain the direct answer. As shown in Figure~\ref{fig:irdata}, collectively the documents may include content to form a summary that can answer the query. This makes our dataset more abstractive in nature as a QMDS model will need to recover and generate the answer passage (summary) using the query and all the other (top-ranked) documents in the triplet.

\section{Models}
\label{models}

\noindent\textbf{Notation.} Our QMDS datasets comprise of instances of triplets of query-documents-summary, $[q,\{D_i\}_i^N,y]$, representing the query $q$, list-of-documents $\{D_i\}_i^N$ and the summary text $y$. Each input document $D_i$ of the triplet, is represented as sequence of tokens, $D_i$=$\{w_{ij}\}_{j=1}^T$, in which $w_{ij}$ is the $j^{th}$ token in the $i^{th}$ ranked document $D_i$. We represent the latent representations as follows: let the input to the encoder of the transformer be $h_{ij}^0$. Then the input and output representations of any transformer encoder block in the $l^{th}$ layer is represented with $h_{ij}^{l-1}$ and $h_{ij}^l$, respectively. 

\begin{figure*}
  \begin{subfigure}[b]{0.5\textwidth}
    \centering
    \includegraphics[width=0.65\textwidth,clip]{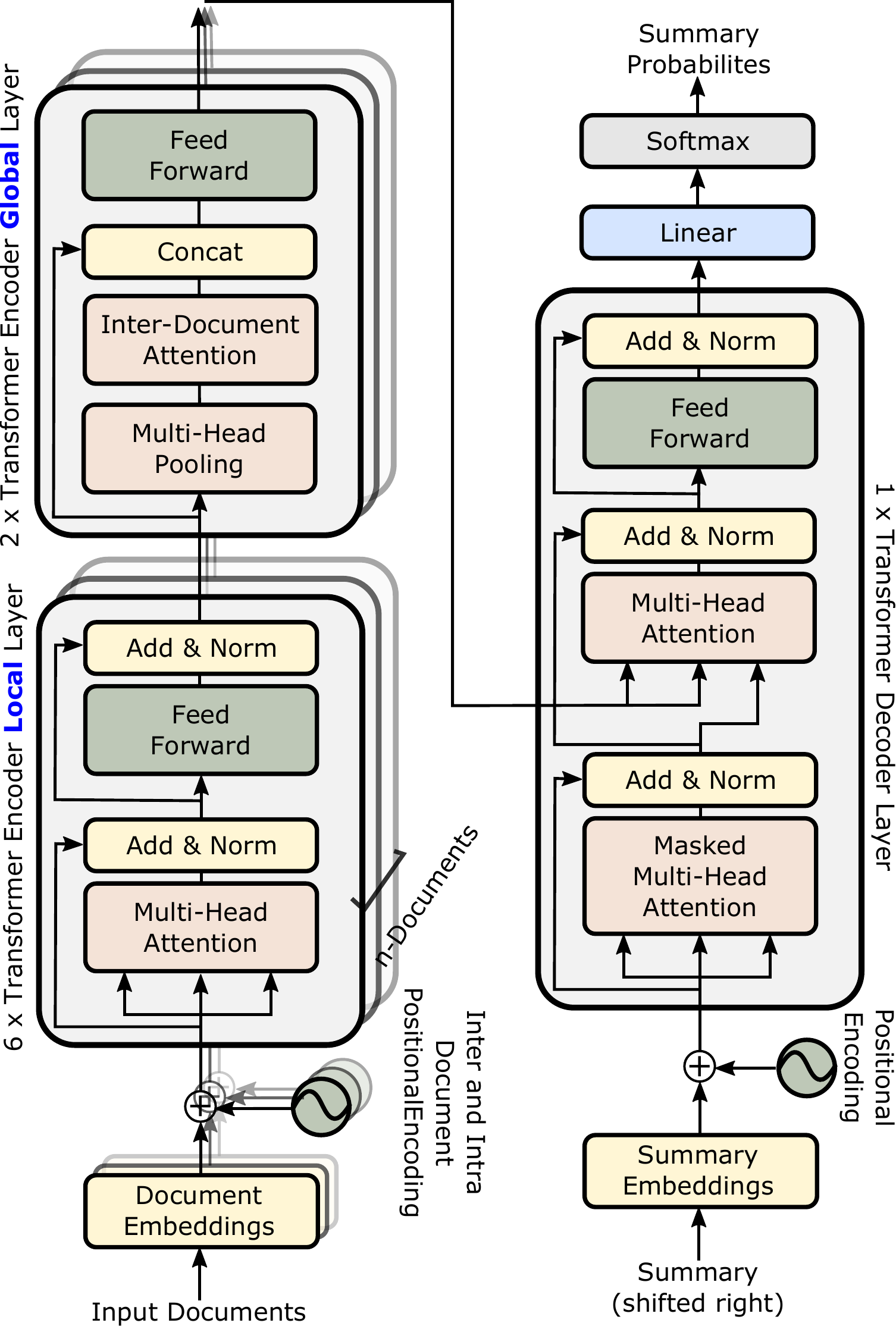}
    \caption{Baseline encoder-decoder QMDS model.}
    \label{fig:qmds-baseline}
  \end{subfigure}
  \begin{subfigure}[b]{0.5\textwidth}
  \centering
    \includegraphics[width=0.7\textwidth]{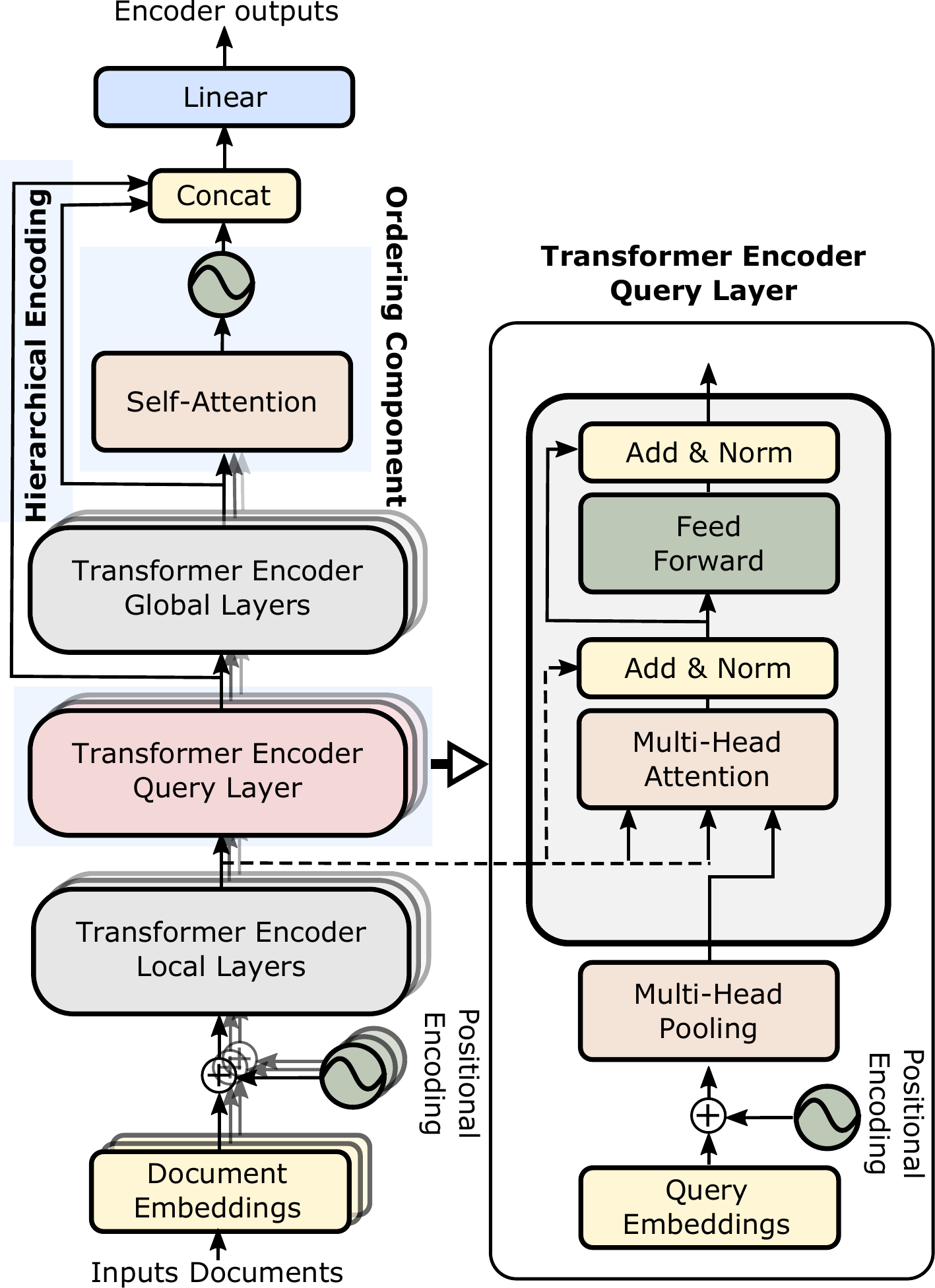}
    \caption{\small The \ourmodel{} encoder.}
    \label{fig:model-components}
  \end{subfigure}
  \caption{Comparison of (a) baseline and (b) \ourmodel{} model with three new components that extends the baseline QMDS model: Hierarchical Encodings, Ordering Component and Query Encoder, enlarged on the right of (b). Unlike baseline model, the \ourmodel{} decoder attends to both the local and global layers.}
\end{figure*}

\subsection{Baseline QMDS Model}

Our baseline is similar to the previous work of~\newcite{liu2019hierarchical},\footnote{\newcite{liu2019hierarchical} considered their model in a MDS setup, however, we view it as a simple QMDS model.} in which multiple documents are first separately encoded via transformer local encoder layers. Next, we add global transformer layers with multi-head pooling and inter-paragraph attention to allow each document to be aware of the information present in the other documents. Later, we use the output of the global encoder as the encoder context for the transformer decoder layers (see Fig.~\ref{fig:qmds-baseline}). In this baseline, we append the query to the first document. Also, we encode the document ranking information in the form of positional encoding which is obtained from a separate document ranker model~\cite{liu2019hierarchical}.

\noindent\textbf{Document Encoding.}
Each word token $w_{ij}$ in each document $D_i$ is mapped to an embedding vector $w^e_{ij}$.
Since transformers have no sequence information, we use position encoding embeddings similar to~\newcite{vaswani2017attention}. Different from SDS models, for MDS we encode both the \textit{inter} and \textit{intra} document positions of each word token $w_{ij}$. For this, we use two positional encoders, one for inter-document level, representing the order of the document and another for intra-document level, representing the position of the token in the document. Then the positional encoding of a token $w_{ij}$ is concatenation of the inter $p^e_{i}$ and intra $p^e_j$ document position encodings, respectively. Finally, the input to the transformer $h_{ij}^0$ is represented as: $h_{ij}^0 = w^e_{ij} + [p^e_i;p^e_j]$, 
where $[;]$ presents the concatenation operation.

\noindent\textbf{Local Transformer Layer.}
We use the same transformer layer proposed in~\newcite{vaswani2017attention} as our local transformer layer. This layer has the traditional multi-head attention module, feed-forward network, and layer normalization. 

\noindent\textbf{Global Transformer Layer.}
Our global transformer layer is similar to that of~\newcite{liu2019hierarchical}, which primarily encodes the inter-document context information. This layer has 3 components: (1) \textbf{multi-head pooling} to obtain a fixed length document representations; (2) \textbf{inter-document attention} to model the dependencies across multiple documents; and (3) \textbf{concatenation of the input} with the context from inter-attention followed by a feed-forward network (see Fig.~\ref{fig:qmds-baseline}). More details on global transformer layer can be found in~\newcite{liu2019hierarchical}.

\noindent\textbf{Decoder Transformer.} We use the same decoder transformer layer proposed in~\newcite{vaswani2017attention}, as shown on the right side of the Fig.~\ref{fig:qmds-baseline}.

\subsection{\ourmodel{} Model}
Extending the baseline model, our \ourmodel{} model introduces three new components: 
a new \textit{encoder} for the query, varying \textit{hierarchical encoding} inputs for the decoder, as well as unsupervised learning of the \textit{order} of the salient concepts to be presented in the generated summary. All these components are targeted on the encoder, so in Fig.~\ref{fig:model-components} we are only showing the encoder part of the model.

\noindent\textbf{Query Encoder.}
Unlike the baseline model in which the query text is simply appended to the top-ranked document before sending it to the encoder, we encode the query via a separate transformer encoder layer. This layer is inserted between local and global layers of the encoder, as shown in  Fig.~\ref{fig:model-components} (with an enlarged view provided on the right of the figure). A separate query layer creates a hierarchy of information encoding, i.e., the local layers enable a rich intra-document feature representation, the query layer conditions this local layer features w.r.t. the given query, and the global layer enable the inter-document feature representation on the query conditioned local layers. 

Let $q_k$ be the $k^{th}$ token in the query, and $h^l_{ij}$ be the output of $j^{th}$ token in $i^{th}$ ranked document of the last local layer before the query layer. The query input representation ($h^q_k$) for the query layer is a combination of its token embeddings ($q^e_k$) and the positional encoding $p^q_k$, which is defined as $h^q_k = w^q_k + p^q_k$.
We encode the query input along with the last local layer output ($h^l_{ij}$) in the following steps to form our transformer encoder query layer:
\begin{equation}
\begin{split}
    o^1_i &= \mathrm{LN}(h^l_i + \mathrm{MHA}(h^l_i, h^l_i, \mathrm{MHP}(h^q))) \\
    o^2_i &= \mathrm{LN}(o^1_i + \mathrm{FFN}(o^1_i))
\end{split}
\end{equation}
where, $\mathrm{MHA}$ is multi-head attention, $\mathrm{MHP}$ is multi-head pooling~\cite{liu2019hierarchical} which is applied on full query tokens ($h^q$), $\mathrm{LN}$ is layer normalization, and $\mathrm{FFN}$ is feed-forward networks. $o^2_i$ is the output from this layer which is used as input to the transformer encoder global layer.

\noindent\textbf{Hierarchical Encodings.}
Unlike the baseline model \citep{liu2019hierarchical}, in which the decoder only attends to the global layer features, the \ourmodel{} decoder attends to both the output of the local and global layers taking into account both context. 
Our intuition is that the local layers carry information specific to the individual documents, while the global layers carry information w.r.t. all the documents. 
Specifically, the decoder utilizes the global properties from all documents by attending over to the output of the global layer. It can also attend to the local layers to focus on the specific aspects of the documents, in which salient information related to the query may be more pronounced.
We concatenate the output of the local and global layers and project it through a linear layer, as shown in Fig.~\ref{fig:model-components} top-left. 

\noindent\textbf{Self Ordering Transformer Encodings.}
In QMDS, the rank-order of the list of documents is an important information as it helps the model to weigh in on the documents relevant to the query. Otherwise, focusing equally on all documents makes it very hard for the model to weed out the salient information and also present them in the correct order in the summary. 
Previous work~\cite{liu2019hierarchical} introduced a two-stage pipeline to inject the ordering into their model. In the stage-1, a document ranker is trained separately to learn the importance of a document w.r.t. a given query. In the stage-2, they use these importance scores to rank the documents and encode them in the model. However, the errors introduced by the document ranker can potentially have cascading effects on the performance of the final summarization model. 

To address this issue, we propose a single-stage model that jointly learns to rank the documents while learning to generate salient summary via our ordering component (see Fig.~\ref{fig:model-components}). Instead of using the positional encoding of the document positions predicted by the document ranker, we ignore the document position embeddings in the initial layer of the transformer, and encode the positional embeddings of the documents at the final layer of the transformer encoder. For this, we use a self-attention module~\cite{lin2017structured} to know the importance of each document in the multi-document setup. We then encode this importance information in a novel way via a positional encoding module: 
\begin{equation}
\begin{split}
    \mathrm{PE}_{(D_i,2j)} &= \mathrm{sin}(r_i/10000^{2j/d_{model}}) \\
    \mathrm{PE}_{(D_i,2j+1)} &= \mathrm{cos}(r_i/10000^{2j/d_{model}})
\end{split}
\end{equation}
where, $\mathrm{PE}(D_i,2j)$ represents the ${2j}^{th}$ dimensional positional encoding representation of document $D_i$, $r_i$ is the importance score assigned for document $D_i$ using the self-attention module, and $d_{model}$ is the model's hidden size. This positional encoding module allows us to convert an unordered importance score into an ordering representation, since unordered score projected on a sinusoidal wave are positioned in an orderly fashion. Finally, we concatenate the final global layer representations of the encoder with the document ordering-based positional encoding representations to form the final encoder representations 
(Fig.~\ref{fig:model-components}, top-left).

\section{Experimental Setup\footnote{We provide additional details on datasets and training settings in Sec.~\ref{datadetails} of the Appendix.}}
\label{sec:experiment-setup}

\noindent\textbf{Datasets.}
\label{subsec:datasets}
We use three large datasets for training QMDS models: our two datasets \newdatasetname{} and \irdatasetname{}, described in Sec.~\ref{sec:cnndm-qmds-dataset}, and the WikiSum. We also use DUC 2006 and DUC 2007 datasets for evaluating our models.\footnote{ \url{https://www-nlpir.nist.gov/projects/duc/data.html}}

\noindent\textbf{Model Ablations and Training.} We experiment with four different ablations of \ourmodel{} (\textbf{HS} in short) model extending the baseline QMDS model of \newcite{liu2019hierarchical} with only hierarchical encodings, only the ordering component, and with the query encoding. \textbf{HS-Joint}, our full model, combines two or three of these components depending on the type of the dataset used in the experiments. 

\noindent\textbf{Evaluation Metrics.}
We use ROUGE~\cite{lin2004rouge}, i.e., ROUGE-1, ROUGE-2, and ROUGE-L 
as our automatic evaluation metrics. We report sentence-level ROUGE F1 scores for all large-scale datasets. For DUC datasets, we report sentence-level ROUGE recall with $250$ word limit.

\section{Results}
\label{results}

\begin{table} [t]
\begin{center}
\small
\begin{tabular}{l|c|c|c}
\toprule
Model & R-1 & R-2 & R-L  \\
\midrule
\newcite{liu2019hierarchical}$^\star$ & 38.03 & 24.68 & 36.20 \\
HS w/ Hierarchical Encodings & 38.14 & 24.88 & 36.33 \\
HS w/ Ordering Component & \textbf{38.57} & \textbf{25.13} & \textbf{36.71} \\
HS w/ Query Encoding & 35.70 & 21.86 & 33.70 \\
HS-Joint Model & 38.37 & 24.90 & 36.52 \\
\bottomrule
\end{tabular}
\end{center}
\caption{Performance of our baseline and variations of \ourmodel{} (HS) model on WikiSum dataset. R-1, R-2, and R-L denote sentence-level ROUGE-1, ROUGE-2, and ROUGE-L, respectively. $\star$ is the reproduced result from the code provided by~\newcite{liu2019hierarchical}.
}
\label{table:wikisum-results}
\end{table}

We present empirical results of our proposed models on various datasets. We first report on three large-scale QMDS augmented datasets: WikiSum, \newdatasetname{}, and \irdatasetname{}, to understand how well various models fit the augmented data. 
Validating the superiority of our proposed models and augmentation methods, we also show transfer results of training on our augmented datasets by using DUC 2006 and 2007 (two human-annotated real QMDS datasets) as test sets.

\subsection{Results on Accuracy}
\noindent\textbf{WikiSum Dataset.} Table~\ref{table:wikisum-results} shows the results on the WikiSum dataset. 
We observe that both hierarchical encodings and ordering methods improve the performance of the model in comparison to the corresponding baseline.\footnote{\ourmodel{} (HS) with hierarchical encodings and HS with ordering method are statistically significantly better than baseline with $p<0.05$ and $p<0.01$, respectively, in all metrics.}\footnote{We initially tried the random ranking order of input documents, and it performed worse than original order (baseline in Table~\ref{table:wikisum-results}), which in turn performed lower than our ordering component.} However, the addition of separate query encoding did not improve the results, in fact, they become worse. This can be explained by the fact that this dataset may not be well suited for evaluating the QMDS models since the queries are constructed from the title of the Wikipedia article while the summaries are taken as the  first paragraph of the article. Thus, neither the queries nor the summaries are natural nor constructed to reflect the properties of a high-quality QMDS dataset. 
Finally, we combine the hierarchical and ordering methods to form the joint model (see Fig.~\ref{table:wikisum-results}) which again performs significantly better than the baseline with $p<0.05$ in all metrics.

\noindent\textbf{\newdatasetname{} Dataset.} Table~\ref{table:cnndm-results} presents the evaluation results of our baseline and three of our \ourmodel{} model variations (using hierarchical encodings, ordering, and query encoding) on the new \newdatasetname{} dataset. 
We observe that both HS models with hierarchical encodings and query-based methods perform significantly better than the baseline, however, HS with ordering method did not work well on this dataset.\footnote{Both hierarchical encodings and query-based methods perform significantly better than baseline with $p{<}0.01$ in all metrics. Ordering method also performed well on ROUGE-1/L ($p{<}0.01$).} For this experiment, our HS-Joint model combines the hierarchical encodings and the query encoder components. We observe that HS-Joint model is significantly better in context match accuracy than the baseline with $p<0.01$. Our HS with hierarchical encodings method outperformed the HS-Joint model. This can be attributed to the fact that hierarchical modeling of local and global information is more crucial for this dataset while summaries don't share complementary information with the query.

\begin{table}[t]
\begin{center}
\small
\begin{tabular}{l|c|c|c}
\toprule
Model & R-1 & R-2 & R-L  \\
\midrule
\newcite{liu2019hierarchical} & 36.31 &	15.40 & 33.38 \\
HS w/ Hierarchical Encodings & \textbf{37.88} & \textbf{16.36} & \textbf{35.23} \\
HS w/ Ordering Component & 36.95 & 14.95 & 34.34 \\
HS w/ Query Encoding & 36.96 & 16.05 & 34.37 \\
HS-Joint Model & 37.09	& 16.33 & 34.45 \\
\bottomrule
\end{tabular}
\end{center}
\caption{Accuracy results on \newdatasetname{} dataset.
}
\label{table:cnndm-results}
\end{table}
\begin{table}
\begin{center}
\small
\begin{tabular}{l|c|c|c}
\toprule
Model & R-1 & R-2 & R-L  \\
\midrule
\newcite{liu2019hierarchical} & 43.60 & 21.88 & 39.40 \\
HS w/ Hierarchical Encodings & 43.37 & 21.64 & 39.21 \\
HS w/ Ordering Component & 39.37 & 18.79 & 35.61 \\
HS w/ Query Encoding & 44.11 & 22.62 & 39.93 \\
HS-Joint Model & \textbf{45.53} & \textbf{23.44} & \textbf{41.15} \\
\bottomrule
\end{tabular}
\end{center}
\caption{Accuracy results on \irdatasetname{} dataset.
}
\label{table:bing-results}
\end{table}

\begin{table*}
\begin{center}
\small
\begin{tabular}{rl|cccc|cccc|cccc}
&\multicolumn{1}{c}{~}
&\multicolumn{4}{c}{(a) {\bf DUC 2006 test set}}
&\multicolumn{4}{c}{(b) {\bf DUC 2007 test set}}
&\multicolumn{4}{c}{(c) {\bf DUC 2007 test set}}\\
\toprule
Model & Dataset & 
R-1 & R-2 & R-L & R-SU4 &
R-1 & R-2 & R-L & R-SU4 &
R-1 & R-2 & R-L & R-SU4 \\
\midrule
\multirow{5}{*}{Baseline} &
No-Pretraining & - & - & - & - 
               & - & - & - & - 
               & 10.01 & 1.42 & 9.80 & 3.18 \\
&WikiSum & 24.00 & 4.28 & 22.72 & 8.16
         & 23.42 & 4.41 & 22.17 & 8.05
         & 34.34 & 6.35 & 32.07 & 11.42 \\
&\irdatasetname{} & 29.65 & 3.83 & 27.93 & 9.63
                  & 29.35 & 3.75 & 27.45 & 9.55
                  & 32.81 & 4.15 & 30.54 & 10.53 \\
&\newdatasetname{} & 30.45 & 6.13 & 28.61 & 10.39
                   & 31.72 & \textbf{7.07} & 29.69 & 11.34
                   & 36.80 & \textbf{7.36} & 34.49 & 12.53 \\
&\jointdatasetname{} & \textbf{30.57} & \textbf{6.17} & \textbf{28.88} & \textbf{10.57}
                     & \textbf{32.33} & 6.98 & \textbf{30.50} & \textbf{11.63}
                     & \textbf{37.07} & \textbf{7.36} & \textbf{34.62} & \textbf{12.60}\\
\midrule
\multirow{5}{*}{HS-Joint} &
No-Pretraining & - & - & - & -
               & - & - & - & -
               & 18.80 & 2.40 & 18.17 & 5.94\\
&WikiSum & 22.96 & 4.09 & 21.76 & 7.89
         & 22.91 & 4.45 & 21.74 & 7.92
         & 29.97 & 4.22 & 27.98 & 9.10 \\
&\irdatasetname{} & 30.17 & 4.01 & 28.31 & 9.79
                  & 29.74 & 3.74 & 27.83 & 9.71
                  & 31.33 & 4.02 & 29.29 & 10.10 \\
&\newdatasetname{} & \textbf{31.14} & \textbf{6.28} & \textbf{29.28} & 10.90
                   & \textbf{34.14} & \textbf{7.60} & \textbf{32.08} & \textbf{12.50}
                   & \textbf{38.31} & \textbf{7.64} & \textbf{35.65} & \textbf{13.26} \\
&\jointdatasetname{} & 30.83 & 6.18 & 29.17 & \textbf{10.91}
                     & 33.13 & 7.37 & 31.05 & 12.04
                     & 36.26 & 6.49 & 33.79 & 12.34 \\
\bottomrule
\end{tabular}
\end{center}
\caption{Transfer results where each model is trained on various datasets, and tested respectively on (a) DUC 2006 and (b) DUC 2007.
In (c), the pre-trained models of (b) are then fine-tuned on DUC 2006 dataset.
}
\label{table:transfer}
\end{table*}

\noindent\textbf{\irdatasetname{} Dataset.} Table~\ref{table:bing-results} shows the results on \irdatasetname{} dataset, comparing our model ablations against the baseline. For this experiment, our HS-Joint model is a combination of hierarchical encodings and query encoding components. We observe that HS with query encodings method performs significantly better than the baseline (with $p<0.01$ on ROUGE-1 and ROUGE-2 metrics, and $p<0.05$ on ROUGE-L), suggesting that this dataset enables efficient use of the queries by the QMDS models. Overall, we achieve best results with our HS-Joint model in comparison to our baseline and other HS ablations with $p$$<$$0.01$ in all metrics. 
%
%
\subsection{Results on Transfer Learning}
We use the DUC 2006 and 2007 datasets for transfer learning experiments with two scenarios.
In the first scenario, we train on the 3 large QMDS datasets and show transfer results on the DUC 2006 and 2007 datasets. In the second one, we finetune models from the first scenario on DUC 2006, and then test on DUC 2007. We evaluate on quantitative and qualitative metrics using ROUGE and human evaluations, respectively. In both scenarios, we compare results of the baseline model~\citep{liu2019hierarchical} to our HS-Joint model, which is the last row in Table~\ref{table:wikisum-results},~\ref{table:cnndm-results}, \&~\ref{table:bing-results}. Our data augmentation methods are not specific to solve DUC datasets, but rather aim to improve QMDS in general, where DUC is one of the standard evaluation sets on which we show improvements via transfer setup. We believe our data augmentation methods would be useful for the community in creating larger-scale training datasets for QMDS. 

\noindent\textbf{Impact of our data augmentation methods.}
Table~\ref{table:transfer}
shows results when DUC 2006 and DUC 2007 datasets are used as test sets and compare our HS-Joint models against the baseline models.
We report recall scores with $250$ word length. Based on pre-training experiment results on DUC 2006 in Table~\ref{table:transfer}(a) and on DUC 2007 in Table~\ref{table:transfer}(b), our data augmentation methods perform better than training on the WikiSum dataset by a large margin.
Our baseline models trained on the combined datasets, \jointdatasetname{}, outperform all other baseline models. However, on the HS-joint models, \jointdatasetname{} is not better than individual data augmentation methods. This suggests that we might also need better weighted sampling or curriculum learning when we combine these two datasets, which we leave for future work.
However, we believe that the individual contributions of our two data augmentation methods (QMDSCNN and QMDSIR) are still useful.
We observe similar behavior on the DUC 2007 after fine-tuning in Table~\ref{table:transfer}(c). 

\noindent\textbf{Impact of new QMDS components.}
Compared to the baseline model, our HS-Joint models, which incorporate novel QMDS focused components, yield much better results when trained on datasets constructed with data augmentation methods and tested on real human datasets as shown in Table~\ref{table:transfer}.
Results support that with better data augmentation and a much better transformer architecture,
we can build more accurate models with higher transfer capabilities.\footnote{The current SOTA extractive QMDS model~\cite{roitman2020unsupervised} achieves R-1/R-2/R-SU4 scores of 43.94/10.09/15.96 on DUC 2006 and 46.02/12.53/17.91 on DUC 2007. However, it is not strictly comparable with our end-to-end abstractive QMDS models.}

\noindent\textbf{Human evaluation.}
We also evaluate our data augmentation methods using head-to-head human evaluations on Amazon Mechanical Turk (AMT). We compare summaries generated by two baseline models: one trained on the WikiSum (WIKI) and another one on the \jointdatasetname{} (CB) dataset, combining our two new datasets. We generate samples from DUC 2006 and DUC 2007 test dataset, and each sample is evaluated by $3$ judges. For DUC 2007, we use the same models fine-tuned on the DUC 2006 dataset as explained earlier. For each test dataset, we ask the turkers to choose between the two model summaries that answer the given query based on $5$ different aspects:\footnote{During AMT evaluation, we also show one of the gold summaries without providing the original documents.}
(1) \emph{Informativeness}: which summary is better in terms of answering the query better? (2) \emph{Non-redundancy}: which summary is better in terms of repeating less of the same ideas? (3) \emph{Coherence}: which summary is better in terms of expressing ideas in the clearest manner fluently? (4) \emph{Order}: which summary is better at presenting the information in the logical order? (5) \emph{Focus}: which summary is better in terms of only sharing the main ideas with no extra superfluous details? We also ask the turkers to compare the summaries on overall quality. We chose the turkers who are located in the USA and UK, have at least $10,000$ approved HITs, and have an approval rate of greater than $98\%$. We pay \$0.5 for a HIT.
The results are as shown in Table~\ref{table:human-evaluation}, where on DUC 2006, our data augmentation CB method yields much better results compared to the one on WikiSum in all aspects.\footnote{Our CB augmentation method is statistically significantly better than WikiSum in all 5 aspects with $p<0.001$ based on bootstrap test~\cite{noreen1989computer,efron1994introduction}.} On DUC 2007, a fine-tuning setup, we still see that our method is better in all aspects except focus.\footnote{Even though our method performed lower on the focus aspect, the difference is 
very low (lowest w.r.t. all other aspects).}
Overall, these human evaluations also suggest that our augmentation methods are better than previous work (WikiSum).

\begin{table}[t]
\begin{center}
\small
\setlength{\tabcolsep}{0.5em}
\begin{tabular}{l|ccc|ccc}
\toprule
\multirow{2}{*}{Criteria} &\multicolumn{3}{c|}{DUC 2006} &\multicolumn{3}{c}{DUC 2007}  \\ \cmidrule{2-7}
 & WIKI & CB & = & WIKI & CB & = \\ 
\midrule
Informativeness & 30 & \textbf{107} & 12 & 52 & \textbf{63} & 20 \\ 
Non-redundancy & 27 & \textbf{110} & 12 & 51 & \textbf{58} & 25 \\
Coherence & 27 & \textbf{112} & 11 & 55 & \textbf{59} & 20 \\
Order & 25 & \textbf{112} & 11 & 56 & \textbf{63} & 12 \\
Focus & 21 & \textbf{117} & 12 & \textbf{61} & 59 & 13 \\
Overall & 23 & \textbf{103} & 24 & 46 & \textbf{53} & 35 \\
\bottomrule
\end{tabular}
\end{center}
\caption{Human evaluation 
between baseline model trained on WikiSum (WIKI) and \jointdatasetname{} (CB) datasets. 
`=' denotes no difference between the two.
}
\label{table:human-evaluation}
\end{table}

\section{Conclusions}
To support research on query-focused multi-document summarization task, we introduce two new data augmentation methods using existing and new data sources. We further introduce a new transformer encoder-decoder model that extends the baseline models with new components to encode the queries together with multiple documents in a hierarchical setting. New components enrich the information provided to the decoder that generates focused summaries. 
We show that summaries generated by the models trained on augmented datasets are more accurate compared to the existing datasets. Additionally, 
our best model can generate summaries that are coherent and contain specific information related to the query with better order of events.

\section*{Acknowledgments}
We thank the reviewers for their helpful comments.
We thank Tong Wang at Microsoft Turing for helping create \irdatasetname{}. We also thank Paul Bennett, Tobias Schnabel, Woon Sang Cho, Chen Qu, and Jiawei Wu for helpful discussions. This work was partially supported by NSF-CAREER Award 1846185 and a Microsoft PhD Fellowship.

\bibliography{citations.bib}

\pagebreak
\appendix
\section{Supplementary Material}
\label{sec:appendix}

\subsection{Additional Details on \newdatasetname{} and \irdatasetname{} Datasets}
\label{newdatadetails}
In this work we presented two data augmentation methods which yielded two new curated datasets, \newdatasetname{} and \irdatasetname{}.  

\paragraph{Is \newdatasetname{} dataset suitable for QMDS? (cont..)} 
Here we provide additional details on our analysis and thought process in evaluating \newdatasetname{} as a good QMDS dataset.

\begin{figure}[ht]
  \begin{subfigure}[b]{0.45\textwidth}
    \includegraphics[width=1.0\textwidth]{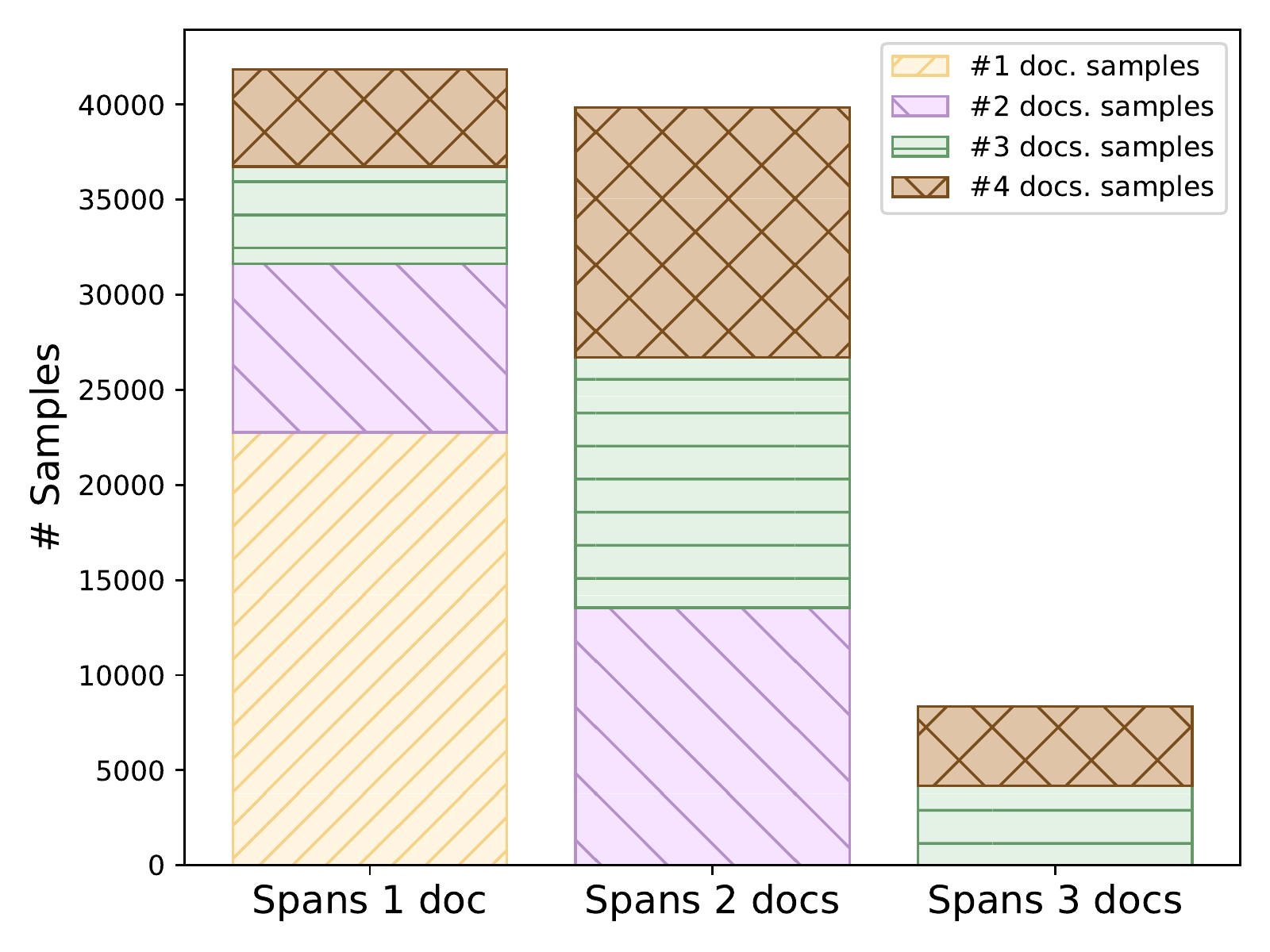}
    \caption{CNN article-summary pairs.}
    \label{fig:cnn-stats}
  \end{subfigure}
  \begin{subfigure}[b]{0.45\textwidth}
    \includegraphics[width=1.0\textwidth]{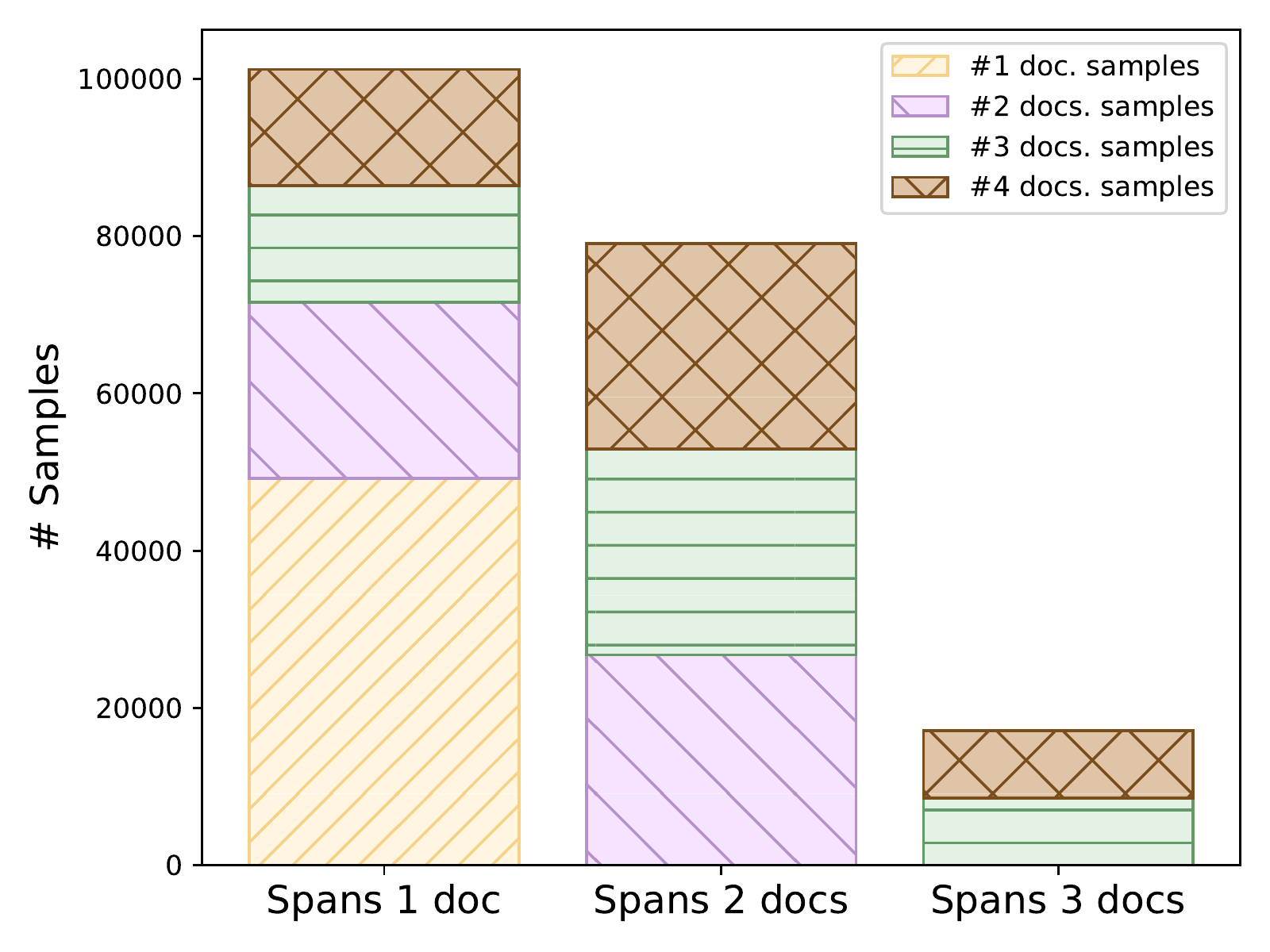}
    \caption{Daily Mail article-summary pairs. 
    }
    \label{fig:dm-stats}
  \end{subfigure}
  \caption{Analysis on CNN and Daily Mail article-summary pairs for multi-document setup. Plot shows the statistics of number of samples whose summary spans $n$ documents. 
   }
\end{figure}

An expected good property of a QMDS dataset that is comprised of query-documents-summary triplets is that, a given summary text should contain sentences that span across multiple input documents. We measure if \newdatasetname{} has this property in the following way: We take a summary and corresponding documents from a query-documents-summary triplet (excluding the retrieved documents), and align each sentence in the summary to one of the documents. 
We first measure the similarity between each summary sentence and each document using ROUGE-L score and map the sentence to the document that has the max-ROUGE-L similarity score. Then, we count the total number of documents that the sentences of the summary are spanning in a query-documents-summary triplet. We call it the alignment score for a given sample. We bin the samples based on this alignment score into $n$ categories, in which a category $k$ contains all the samples whose summary spans $k$ documents. Fig.~\ref{fig:cnn-stats} and Fig.~\ref{fig:dm-stats} show such statistics on CNN and DM samples of our multi-document setup. We can conclude from these figures that there are a good number of samples in our MDS setup whose summary spans multiple documents, thus, enabling a multi-document setup using single document datasets.

\paragraph{\irdatasetname{} Dataset Construction.}
We filter the collected search query logs data making sure the following criteria are met: (\textit{i}) The generated extractive summary should contain at least $2$ sentences. (\textit{ii}) The document list should have at least $3$ retrieved-documents. (\textit{iii}) Each sentence of the summary should be covered by at least one of the input document. We measure the coverage of a summary sentence by a document based on ROUGE-1 matching score. We set a threshold of $0.8$ on the matching score to consider if a summary sentence is covered by a document. This ensures that the information in the target summary passage is included in the corresponding document set to some degree.
This yields us with a total of $102,595$ sample triplets out of which we use $82,076$ for training, $10,259$ for validation, and $10,260$ for testing.

Samples from \newdatasetname{} and \irdatasetname{} Datasets are shown in Table~\ref{tab:samplemultidocuments}.

\begin{table*}
\small
    \centering
    \begin{tabular}{|p{10cm}|p{5cm}|}
        \midrule
        \multicolumn{2}{|l|}{\textcolor{blue}{\textbf{\irdatasetname{}} Dataset}} \\
        \midrule
        \colorbox{yellow}{\textbf{Query} : rock roll hall of fame 2019}
        
        \textbf{Document 1}: [$\dots$] industry vote \textcolor{red}{official ballots are sent to more than 1,000 voters , including every living rock hall inductee , artists , members of the music industry and music historians} . [$\dots$]

        \textbf{Document 2}: [$\dots$] \textcolor{blue}{the 34th annual rock \& roll hall of fame induction ceremony , presented by klipsch audio , will take place on friday , march 29 , 2019 at barclays center in brooklyn , new york }. [$\dots$]
        
\textbf{Document 3}: [$\dots$] news rock and roll hall of fame : 8 best moments from the 2019 induction ceremony 3 / 30 / 2019 by joe lynch inductee robert smith of the cure performs at the 2019 rock \& roll hall of fame induction ceremony at barclays center on march 29 , 2019 in new york city . [$\dots$]

\textbf{Document 4}: [$\dots$] all of the respected artists are slated to attend the lavish ceremony , \textcolor{magenta}{held at barclays center} in brooklyn , although the rock and roll hall of fame foundation noted some members of radiohead will likely be absent . [$\dots$]

\textbf{Document 5}: [$\dots$] com new york , new york \- \- the rock \& roll hall of fame inducted its seven newest members on friday night at the barclays center in brooklyn , n.y . [$\dots$]

         & \textbf{Summary:} \textcolor{blue}{the rock \& roll hall of fame 2019 induction ceremony , presented by klipsch audio , will be} \textcolor{magenta}{held at barclays center} in brooklyn , new york on march 29 , 2019 . ticket on - sale information will be announced in january . \textcolor{red}{ballots will be sent to an international voting body of more than 1 , 000 artists , historians and members of the music industry} . \\
    \midrule
    \multicolumn{2}{|l|}{\textcolor{blue}{\textbf{\newdatasetname{}{}} Dataset}} \\
    \midrule
    \colorbox{yellow}{\textbf{Query}: national hurricane center director leaves position}
    
    \textbf{Document 1}: [$\dots$] \textcolor{blue}{national hurricane center director bill proenza left his position monday} , just days after \textcolor{magenta}{nearly half of the nhc staff signed a petition calling for his ouster .} hurricane center bill proenza left his job as director monday . [$\dots$]

    \textbf{Document 2}: [$\dots$] \textcolor{red}{proenza caused an uproar last month with comments about a key hurricane satellite called quikscat .} the satellite is five years beyond its life expectancy and operating on a backup transmitter . [$\dots$]
    
    \textbf{Document 3}: [$\dots$] \textcolor{magenta}{his staffers on thursday issued a petition calling for him to step down .} watch how proenza lost the confidence of his staff '' the petition said [$\dots$] hurricane center staffers told cnn 's john zarella they were unhappy not only about his comments about the quikscat , but also \textcolor{orange}{about the environment at the center -- one characterized by closed doors and the public airing of dirty laundry .} [$\dots$]
    
    \textbf{Document 4}: [$\dots$] proenza on friday told cnn he had contacted his superiors in washington about `` ways to move forward , '' but added , `` i am not going to resign . [$\dots$]
    
    \textbf{Document 5}: [$\dots$] \textcolor{magenta}{about half of the staff of the national hurricane center have signed a petition calling for the ouster of the center 's director} , saying its `` effective functioning '' is at stake as the atlantic hurricane season heads toward its peak . [$\dots$] the center 's current director , bill proenza , took over in january after the retirement of max mayfield . \textcolor{red}{proenza caused an uproar last month with comments about a key hurricane satellite called quikscat .} [$\dots$]
    
    \textbf{Document 6}: [$\dots$] hurricane hound uses google maps as its framework and tracks both forecasts and the locations of atlantic and east pacific hurricanes and tropical storms  [$\dots$]
    
    \textbf{Document 7}: [$\dots$] from a hurricane readiness viewpoint , one must prepare every season as if a major hurricane will impact them , '' said bill read , director of the national hurricane center . [$\dots$]
    
    \textbf{Document 8}: [$\dots$] the declaration gives the director of the governor 's office of emergency preparedness authority `` to undertake any activity authorized by law which he deems necessary and appropriate '' to prepare for the possibility of a hurricane . [$\dots$]
    
    & \textbf{Summary:} \textcolor{blue}{national hurricane center director bill proenza has left his position .} \textcolor{magenta}{nearly half of the nhc signed a petition calling for him to step down .} \textcolor{red}{proenza came under fire for comments he made about the quikscat satellite .} \textcolor{orange}{staff unhappy with environment of closed doors , public bickering .}  \\
    
    \midrule
    \end{tabular}
    \caption{Example \{Query, Documents, Summary\} triplets from \irdatasetname{} and \newdatasetname{} datasets. }
    \label{tab:samplemultidocuments}
\end{table*}

\begin{figure*}[t]
\centering
\includegraphics[width=0.98\linewidth]{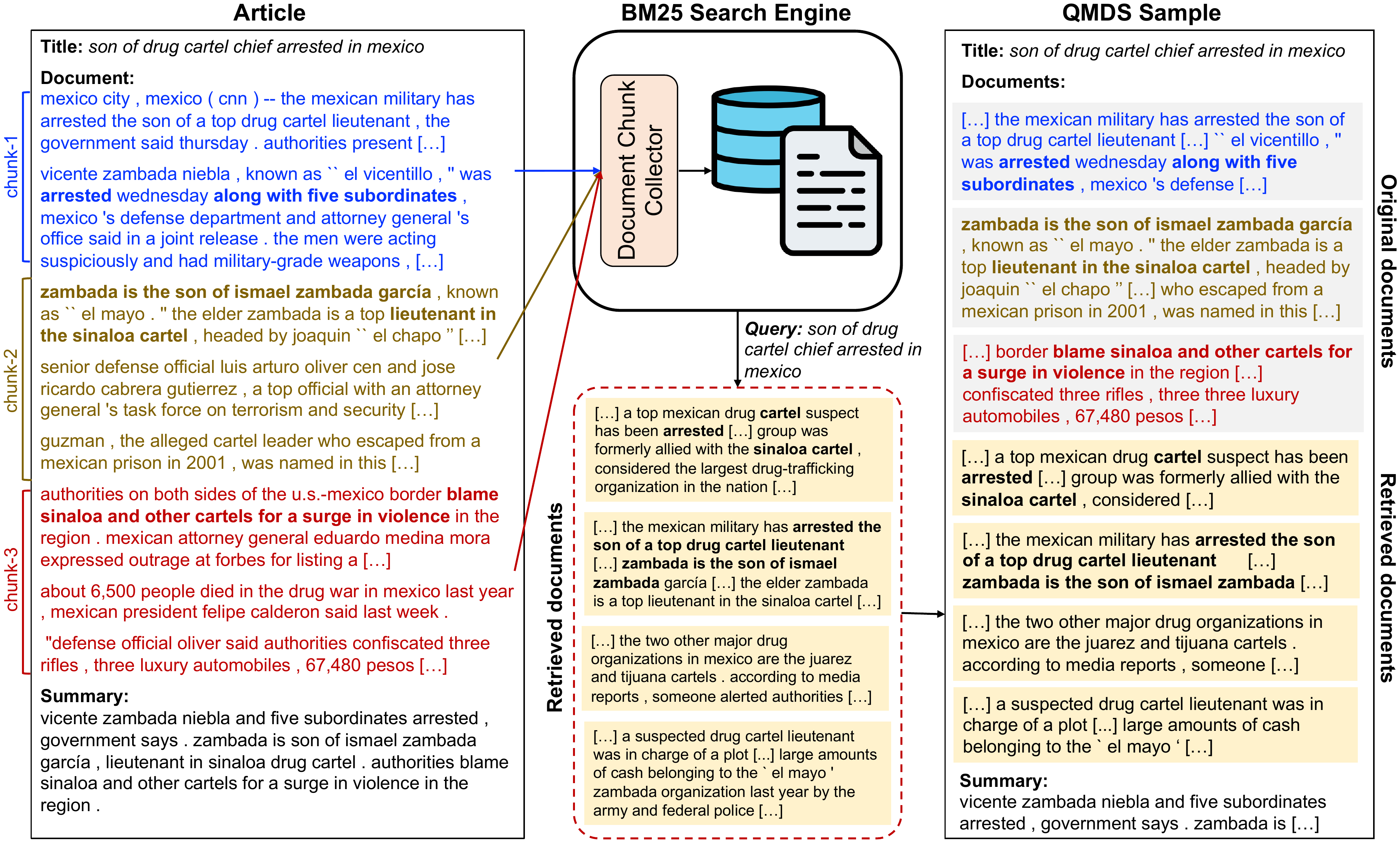}
\caption{High level description of our \newdatasetname{} dataset curation pipeline. Bold text in the documents aligns with the summary content.  
}
\label{fig:dataet-pipeline}
\end{figure*}

%
\subsection{Query Impact on \newdatasetname{} dataset}
\begin{table}
\begin{center}
\small
\begin{tabular}{l|c|c|c}
\toprule
Query & R-1 & R-2 & R-L  \\
\midrule
Original title & \textbf{30.64} & \textbf{10.92} & \textbf{27.88}  \\
Distractor Query & 30.40 & 10.72 & 27.61 \\
Dull Query & 29.37 & 10.21 & 26.79 \\
Dissimilar Query & 29.42 & 10.05 & 26.91 \\
\bottomrule
\end{tabular}
\end{center}
\caption{Effect of various types of queries on the model's performance on the new \newdatasetname{} dataset. 
}
\label{table:cnndm-query-analysis}
\end{table}
When constructing the \newdatasetname{} dataset, we investigated different query types. Here, we analyze the impact of each query type on the performance of the QMDS tasks. Among different query types are: original title, which is the title of the document as query; a distractor query, which is the title of the closest CNN/DM sample ranked based on the current sample's title as query, a dull query such as \textit{what is it?}; and dissimilar query, which is the title of a sample whose ROUGE-1 score with the current sample's title is less than 0.2 on F1 score.
Table~\ref{table:cnndm-query-analysis} shows the baseline models trained on \newdatasetname{} dataset with varying query types. The results are on $500$ samples from the test set.
Using a distractor query reduces the performance, while a dull or dissimilar query adversely affects the performance. These observations suggest that when constructing QMDS datasets, samples with queries that are strongly related with the summary should be included.

\subsection{Experiment Details}
\label{datadetails}

\paragraph{Additional Details on WikiSum Dataset.}

WikiSum dataset~\cite{liu2018generating} is based on Wikipedia articles and the source reference documents. The input is a Wikipedia topic and a collection of non-Wikipedia documents referenced in the Wikipedia article, where the text in the article itself is the output/target. While this dataset is primarily used for MDS, we consider the weak signal from the topic information of the article as query and use it as QMDS dataset in all our experiments. We use the pre-processed publicly available version of this dataset~\cite{liu2019hierarchical}.\footnote{\url{https://github.com/nlpyang/hiersumm}} Overall, this dataset has $1,579,360$ samples for training, $38,144$ for validation, and $38,205$ for testing. The pre-processed dataset has top $40$ documents for each sample which are ranked by~\newcite{liu2019hierarchical}, and the text are encoded via subword tokenization with SentencePiece~\cite{kudo2018sentencepiece}.

\paragraph{Additional Details on DUC 2006 and 2007.} These datasets are introduced as part of 
the Document Understanding Conference (DUC) for QMDS model evaluations.\footnote{\url{https://www-nlpir.nist.gov/projects/duc/data.html}} Each dataset includes topic related document samples and is described as follows: Given a topic and a set of 25 relevant documents, the task is to generate a summary of the documents that answers the questions about the given topic. For the DUC 2006 dataset, we have 50 samples and for the DUC 2007 dataset we have 45 samples.

\paragraph{Model Training Details.}
\label{trainingdetails}
We tune all the hyperparameters on validation sets and select the best checkpoint based on best ROUGE-L validation score. However, we do not perform any extensive hyperparameter search. We report results based on single runs and use the same seed for parameter initialization across various models. We manually tune only learning rate in the range $[1,4]$ and the number of layers in the transformer encoder in the range $[5,8]$ for very few initial experiments.
In all our models, we use $8$ transformer layers\footnote{We use $5$ transformer local layers, $1$ transformer query layer, and $2$ transformer global layers on the encoder side for all models with a separate query. For the rest of the models, we use $6$ transformer local layers and $2$ transformer global layers for the encoder.} on the encoder side and one transformer layer on the decoder side. We set the word embedding size and hidden size of the transformer to $256$, and the feed-forward hidden size to $1024$. We use Adam optimizer with a learning rate of $1$ and enable initial warmup steps of $8000$, and we set dropout to $0.1$. All the parameters are initialized with Kaiming uniform initialization~\cite{he2015delving}. All our models are trained on 4 Nvidia Tesla P100 GPUs for $500,000$ steps. We use a gradient accumulation of 2 steps for \newdatasetname{} models, and 4 steps for the rest. The number of parameters in the baseline model is $14.8$ million, whereas the number of parameters in our HS model with hierarchical encodings, ordering method, and query encoding method are $14.9$, $15.0$, and $15.7$ million, respectively. During the decoding phase, we use beam search with beam size $5$ and a length penalty with $\alpha=0.4$~\cite{wu2016google}.

For training models on the WikiSum dataset, we use a maximum token length of $200$ for each input document and $100$ for the target summary, and limit the number of input documents to $24$, leading to a batch size of $10,500$. The average training time on this dataset is $5$ days. During the testing, we use $40$ documents for the input and set a maximum decoding length of $250$ and a minimum decoding length of $20$.

For training models on our \newdatasetname{} dataset, we use a maximum token length of $200$ for each input document and $100$ for the target summary, and limit the number of input documents to $8$, leading to a batch size of $8,000$. The average training time on this dataset is $2.2$ days. During the testing, we use $8$ documents for the input and set a maximum decoding length of $250$ and a minimum decoding length of $35$. Further, we also discourage trigram repetitions.

For training models on our \irdatasetname{} dataset, we use a maximum token length of $400$ for each input document and $100$ for the target summary, and limit the number of input documents to $6$, leading to a batch size of $7,200$. The average training time on this dataset is $3.5$ days. During the testing, we use $8$ documents for the input and set a maximum decoding length of $400$ and a minimum decoding length of $20$.

During the testing of all models on DUC 2006 and DUC 2007 test sets, use a maximum token length of $800$ for each input document, limit the number of input documents to $25$, and set a minimum decoding length of $400$. Further, we discourage trigram repetitions and set the beam size to $15$. For fine-tuning models on the DUC 2006 dataset, we use a maximum token length of $600$ for each input document and $400$ for the target summary, and limit the number of input documents to $20$, leading to a batch size of $8,000$. For these experiments, we set the gradient accumulation to $8$ steps.

\end{document}